# Acquiring Knowledge for Evaluation of Teachers' Performance in Higher Education – using a Questionnaire

Hafeez Ullah Amin
Institute of Information Technology
Kohat University of Science & Technology (KUST)
Kohat, Pakistan
Email: hafeezullahamin@gmail.com

Abdur Rashid Khan
Institute of Computing and Information Technology
Gomal University, D.I.Khan, Pakistan
Email: rashidkh08@yahoo.com

*Abstract*— in this paper, we present the step by step knowledge acquisition process by choosing a structured method through using a questionnaire as a knowledge acquisition tool. Here we want to depict the problem domain as, "how to evaluate teacher's performance in higher education through the use of expert system technology". The problem is how to acquire the specific knowledge for a selected problem efficiently and effectively from human experts and to encode it in the suitable computer format. Acquiring knowledge from human experts in the process of Expert Systems development is one of the most common problems cited till yet. This questionnaire was sent to 87 domain experts within all public and private universities of Pakistan. Among them 25 domain experts sent their valuable opinions. Most of the domain experts were highly qualified, well experienced and highly responsible persons. The whole questionnaire was divided into 15 main groups of factors, which were further divided into 99 individual questions. These facts were analyzed further to reach a final shape of the questionnaire. This knowledge acquisition technique may be used as a learning tool for further research work.

## I. INTRODUCTION

Expert systems were developed by Artificial Intelligence community as early as in mid-1960s [14]. These systems are decision making and problem solving tools with the abilities to emulate the reasoning of human expert with in a special subject area [1]. The use and development of the expert systems in various subject areas has been found, like agriculture, chemistry, computer science, engineering, geology, medicine, space technology etc. [3]. The critical component in the development of expert system is the knowledge base [7]. This component consists of human experts' knowledge which is extracted by the knowledge engineers through the process called Knowledge Acquisition. The knowledge acquisition is the accumulation, transfer, and transformation of problem solving expertise from human experts to computer program [14]. Acquiring knowledge from experts is difficult task and has been identified as a bottleneck in expert system development [1, 11, 12, 14, 15]. It is among the most common cited problem in the construction of expert system.

The literature study shows that there exist a number of knowledge acquisition techniques for acquiring knowledge and expertise from domain experts, like interview, questionnaire, observation analysis, case study analysis, protocol analysis etc [1, 6, 11, 12, 15]. Further, the selection of a particular technique is based on the target problem to be solved, availability and accessibility of the human experts, and the knowledge engineers' approach towards the solution. Therefore, the problem should clearly be defined; the availability and accessibility of the domain experts should be ensured, and the knowledge engineers should be capable to select a suitable tool for capturing the expertise from concern sources.

In this paper, we present a step by step process of knowledge acquisition through the use of questionnaire for acquiring knowledge from knowledge sources (human experts) in the selected problem that is "evaluating teachers' performance for maintaining teaching quality in higher education". This research work is limited to the higher education institutions in Pakistan, domain experts were selected from universities in Pakistani, and a questionnaire was used as a tool for capturing knowledge.

## II. KNOWLEDGE ACQUISITIONS

Every expert system development begins with the knowledge acquisition process; it is the basic building step to expert system creation. In this process the developers of expert system often called knowledge engineers working with domain experts−(experts who posses expertise in specific subject area), and other sources of knowledge, to collect knowledge in to computer format. The literature describing the knowledge acquisition is a greatest bottleneck in the expert system development process. There are various reasons due to which it becomes such a difficult activity. Ref [11] and [14] described a set of problems that may occur during the process; the major difficulties are:

- Problem is not fully defined.

- Knowledge engineers have lack of knowledge about problem background.
- Maximum number of participants involved.
- Expert may lack of time or unwilling to cooperate.
- Method of knowledge acquisition may be poorly defined.
- Problematic interpersonal communication factors may exist between the knowledge engineers and the experts, etc.

### A. Knowledge Acquisition Techniques

Knowledge acquisition techniques are the methods through which expertise can be captured from knowledge sources, these sources may be human experts, books, journals, data bases, reports, others computer system. Literature study reveals that there exist a number of knowledge acquisition techniques; Interview, observation analysis, protocol analysis, case study analysis, questionnaire, simulation and prototyping [1, 12, 15]. These techniques can be used according to the situation, need, nature of problem, and domain experts. If the domain experts are easily accessible, then interview is the best technique. If the expert do not have free time and are far away from access of knowledge engineers then questionnaire may be used. Similarly, if the target problem is complex, multiple techniques can be used.

### III. APPROACH

Problem identification, selection of domain experts, selection of knowledge acquisition technique, designing a questionnaire, analysis the experts' opinions, and assigning weights to the evaluating criteria, are the important steps of knowledge acquisition process. The whole knowledge acquisition process detail is following:

### A. Problem Identification

The first step in knowledge acquisition process was to identify the critical factors that affect the teachers' performance directly or indirectly. Blank ACR (Annual Confidential Reports) of various institutions, criteria of various awards designed by HEC (Higher Education Commission), Pakistan, and views of various researchers were given due consideration during development of the questionnaire.

Literature study reveals that research has been conducted on different aspects of employees' performance efficacy. As teachers are a type of workforce in educational institutions, but with different jobs description from others employee; thus there should be different criteria for their evaluation. A number of research works conducted in quality of teachers, teachers' assessment, teachers' training etc. Ref [10] described that quality of teacher is a key in any education system, he also highlighted the reasons behind the low quality of teaching that are, low educational qualification, teaching practices, non-existence of proper monitoring system or effective supervision. Ref [9] stated that teacher must contain aspects like effectiveness in teaching, teaching methodology, educational psychology, use of audio-visual aid, evaluation techniques etc.

Ref [5] has worked on the use of student achievement scores as basis for assessing teachers' instructional effectiveness; test scores of students are used as a measure of not only student achievement, but also of teacher achievement, performance and effectiveness. HR (Human Resource) practices (selection, training, promotion, pay etc.) have a direct impact on employee skills, motivation, job design and work structures, also these variables draw out certain levels of creativity, productivity and discretionary effort, which has an impact on profitability and growth, which have a direct impact on the organization' market value [16]. There are positive relationships between compensation, promotion, and evaluation practices impacting employees' and university teachers' performance [2, 13]. Motivation theories raised needs and satisfaction of individual worker, including needs such like food, drink, sleep, and sex, and on a higher plane, things such as security, love, status, recognition, self-respect, growth, accomplishment etc. factors such as supervision, relationship with supervisor, work conditions, salary, company policy and administration all impacting the organization climate and individual performance [4,8]. Since teachers are a type of employees, who also require motivations to teach better, thus these theories are also considered for preparing teachers evaluation criteria. Thus a comprehensive set of attributes have been extracted from literature review, which have been cited as important to the workforce performance in organizations.

### B. Selection of Domain Experts

After determination of the target problem's characteristics and specification domain knowledge areas, the next task was to choose the domain experts who posses a vast knowledge and expertise of the problem domain. In this step the questionnaire was evaluated by domain experts so that it may become a framework for assessment of teachers' performance evaluation. The most important and critical component of the expert systems is the knowledge base: collection of expert knowledge; the human experts are the main source of this knowledge [7]. Due to the nature of the problem domain, experts for selection of decision making parameters were chosen from the fields of education, psychology, and human resource management. That is why professors, senior academicians, and researchers in the field of education, psychology, and human resource were chosen as domain experts for evaluation of the critical factors for decision making. These experts were traced in the 124 (67 public sectors and 57 private) universities of Pakistan. We found 102 domain relative experts along with other relevant information about their qualification, contact address, area of interest, and work experience. Table I, depicts the characteristics of the domain experts.

### C. Selection of Knowledge Acquisition Technique

To acquire knowledge from domain experts, the knowledge engineers needs some techniques to use, which works as tool and support the knowledge engineers in the process of knowledge acquisition. Literature study reveals that there are number of techniques exist for acquiring knowledge from domain experts like interview, questionnaire, protocol analysis, psychological scaling, and card sorting etc [1,11,12,15].

TABLE I. CHARACTERISTICS OF DOMAIN EXPERTS

| Domain Experts' Characteristics |
|---|
| **Gender** |
| – Male |
| – Female |
| **Designation** |
| – Professor |
| – Associate Professor |
| – Assistant Professor |
| – Lecturer |
| **Qualification** |
| – Post Doc |
| – PhD |
| – M.Phil |
| **Administrative Experience** |
| – Vice Chancellor |
| – Dean |
| – Chairman |
| – Director |
| **Subject Specialty** |
| – Education |
| – Human Resource Management |
| – Psychology |
| – Computer Science |
| – Statistics |
| **Demographics** |
| – Federal |
| – Sindh |
| – Punjab |
| – NWFP |
| – Balochistan |

Selection of these experts is based upon; the nature of target problem, skills and experience of knowledge engineers, availability and accessibility of domain experts. Details study of each and every technique is beyond the scope of this paper; however many articles are available in literature of knowledge acquisition techniques. By determining the nature of the problem, its major characteristics, availability and accessibility of domain experts, it was decided to use questionnaire as a tool for acquiring knowledge.

### D. Design and Distribution of Questionnaire

The purpose of this questionnaire is to solicit experts' opinion on given criteria that have been extracted from different documented sources to develop standard criteria which posses all the factors that can affect teachers' performance in higher education institutions. A questionnaire was designed with **15** groups of factors for evaluating teachers' performance. These factors were extracted from different sources as discussed in section 3. Initially there were **107** sub factors relevant to the main factors. As almost the entire questionnaire contain qualitative attributes, which are difficult to measure as like numeric variables' value. Therefore, a technique developed by Zadeh, [17] called fuzzy logic was used. Five fuzzy variables scales have been defined as: 5= critically important, 4= important, 3= helpful, 2= minimally affects, 1= do not affect teachers' performance, which represents a gradual transition from high to low affects on teachers' performance by the various factors in questionnaire (See APPENDIX-I for details).

The questionnaire was designed in such a way that the respondents have to rank the importance of each factor regarding its objective. All the respondents were asked to give suggestions about to change existing factors, delete irrelevant factors, insert new factors, and move one sub factor from one main factor to other as per their importance and relevancy. This ranking of factors was also critical in determining the weights for the various factors influencing the decision making process for individual teacher's performance. Total **87** questionnaires were distributed among these experts due to their availability. Table II, depicts the summary of questionnaire distribution. A maximum of two months time was allowed to send their expert opinions.

### E. Analysis and findings

Only from **25** valuable experts opinions were received well in time (see Table II). After receiving the experts' responses, many experts were again contacted to clarify the respondents' comments and fully elicit the available expertise. Due consideration was given to expert opinions and a questionnaire of questions was finally designed. The experts' modification and their comments were considered and a final set of **15** groups of factors with **99** sub factors was prepared.

Statistical analysis presents that personal abilities and teaching learning process are the most important among the rest of the criteria for evaluating teachers' performance, the administrative skills, responsibilities & punctuality factors, compensation & rewards factors, and the job security & Environment factors showed a nearly similar importance. Promotion factors, research, professional ethics, and supervision also resulted in group of high values in teaching performance. The achievement, organization evaluation policy, needs, and background factors showed enough affects but lower than the others top factors. **Table III** presents the summarized statistical results. Domain experts' knowledge and heuristic approach was used during priority assignments to decision making factors, which have been shown in TABLE IV in sorted order.

TABLE II. QUESTIONNAIRE DISTRIBUTION SUMMARY

| Medium | Questionnaire sent | Responses | %age |
|---|---|---|---|
| Through Post | 50 | 10 | 20.00 |
| Through Email | 21 | 6 | 28.57 |
| By hand | 16 | 9 | 56.25 |
| **Total** | **87** | **25** | **28.73** |

## IV. CONCLUSIONS AND FUTURE RECOMMENDATION

A complete knowledge acquisition process has been adapted to the problem for evaluating teachers' performance in higher education by using the questionnaire as a knowledge extraction tool, as shown in **Fig. 1**. A set of criteria have been developed for the evaluation of teachers' performance through this knowledge acquisition process. This criterion will be used as a base for final decision making. A summation function will be used to calculate the overall score at run time to get the result for ranking. It is also concluded that a questionnaire is the best tool for knowledge acquisition from multiple experts located in diverse locations and if difficult to access directly.

This approach may be used in other similar problems solutions like courses selection, supervisor selection, courses design, and employee assessment.

TABLE III. RESPONSE SUMMARY FOR AFFECT ON TEACHERS' PERFORMANCE
*(Total Reponses N: 25, 5= Critical Affects, 1=Do not Affect)*

| Main Groups of Factors | Mean | Standard Deviation |
|---|---|---|
| Personal Abilities | 4.48 | 0.653 |
| Teaching Learning Process | 4.36 | 0.810 |
| Responsibility & Punctuality | 4.28 | 0.936 |
| Administrative Skills | 4.28 | 0.737 |
| Supervision | 4.12 | 1.053 |
| Professional Ethics | 4.16 | 0.986 |
| Research Orientation | 4.32 | 1.029 |
| Publication | 4.00 | 1.080 |
| Awards & Achievements | 3.96 | 0.934 |
| Compensation & Rewards | 4.28 | 0.842 |
| Promotion Factors | 4.08 | 0.862 |
| Job Security & Environment Factors | 4.28 | 0.842 |
| Organization Evaluation Policy | 3.96 | 1.059 |
| Needs & Requirements | 3.88 | 1.235 |
| Background Factors | 3.72 | 1.275 |

TABLE IV. WEIGHT ASSIGNMENT TO MAIN FACTORS

| S.No | Main Groups of Factors | Weights |
|---|---|---|
| 1 | Research Orientation | 0.0753 |
| 2 | Publication | 0.0742 |
| 3 | Teaching Learning Process | 0.0729 |
| 4 | Personal Abilities | 0.0727 |
| 5 | Responsibility & Punctuality | 0.0726 |
| 6 | Compensation & Rewards | 0.0726 |
| 7 | Professional Ethics | 0.0720 |
| 8 | Job Security & Environment Factors | 0.0706 |
| 9 | Supervision | 0.0677 |
| 10 | Administrative Skills | 0.0674 |
| 11 | Awards & Achievements | 0.0605 |
| 12 | Promotion Factors | 0.0602 |
| 13 | Organization Evaluation Policy | 0.0577 |
| 14 | Needs & Requirements | 0.0550 |
| 15 | Background Factors | 0.0490 |
| | **Total weight:** | **1.0000** |

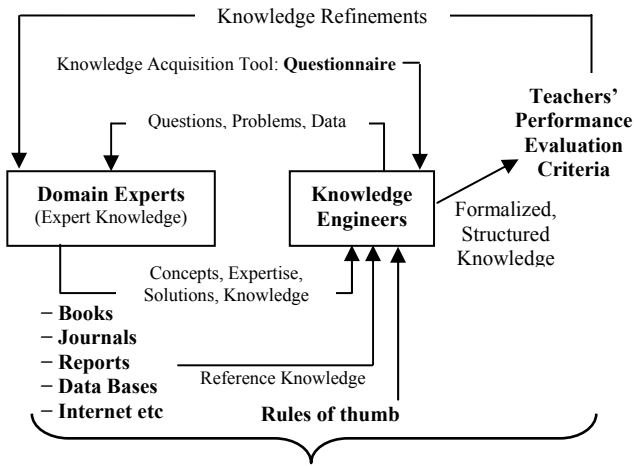

Figure 1. Knowledge Acquisition Process

THE AUTHORS

H. Amin

The author is currently pursuing his MS degree in Computer Science from the Institute of Information Technology, Kohat University of Science & Technology (KUST), Kohat, Pakistan. This work is the part of his MS thesis research work. His area of interest includes expert system, applications of fuzzy logic, databases and data mining.

Abdur Rashid Khan

The author is presently working as an Associate Professor at ICIT, Gomal University D.I.Khan, Pakistan. He received his PhD degree from Kyrgyz Republic in 2004. He has been published more than 16 research papers in national and international journals and conferences with a monograph and a software patent. His research interest includes AI, MIS, Software Engineering, and Data Bases.

**APPENDIX-I**

**A SURVEY ON FACTORS AFFECTING TEACHER'S PERFORMANCE IN HIGHER EDUCATION**

**Respondent's Name:** ___________________ **Designation:** ___________________
**Address/University:** ___________________
**Qualification:** ___________________ **Gender:** ___________________
**Age:** ___________________ **Email:** ___________________

### INSTRUCTIONS

Different aspects of teacher's performance which depends upon attributes like Personal, Teaching, Responsibility & Punctuality and Administrative Attributes etc are included in the scale. As an **_EXPERT_** in this area, you are requested to examine each item in terms of suitability and then to explain the degree of your agreement to each item whether, in your opinion, it would measure the factors affecting teachers' performance and to what extent. You may recommend new & delete unnecessary items from the existing scale. Your in-time response will highly be appreciated. Please, use the scale below to mark (✓) your responses in the area provided.

| 5 | 4 | 3 | 2 | 1 |
|---|---|---|---|---|
| **Critical** to Teacher's Performance | **Important** to Teacher's Performance | **Helpful** to Teacher's performance | **Minimally** affects teacher's performance | **Do not** affect teacher's performance |

| | ATTRIBUTES | | LEVELS | | | | |
|---|---|---|---|---|---|---|---|
| | | | 5 | 4 | 3 | 2 | 1 |
| **1** | **PERSONAL ABILITIES** | | | | | | |
| 1.1 | **Intellectual Ability:** | (The power of reasoning, thinking, and understanding) | | | | | |
| 1.2 | **Analytical Skills:** | (The power of analysis) | | | | | |
| 1.3 | **Creativity:** | (capable to create/ The quality of doing in a new way or new ideas) | | | | | |
| 1.4 | **Maturity :** | (The quality of thinking and behaving in an appropriate manner) | | | | | |
| 1.5 | **Integrity :** | ( Syn: Unity: The quality of being and strong moral principals) | | | | | |
| 1.6 | **Self Confidence:** | (The internal feeling of certainty in oneself) | | | | | |
| 1.7 | **Problem Solving Skills:** | (The ability to solve problems) | | | | | |
| 1.8 | **Cooperative:** | (The ability to work jointly) | | | | | |
| 1.9 | **Intelligence:** | (The ability of learning, understanding in a logical way) | | | | | |
| 1.10 | **Reliability and Dependability:** | (The qualities that can be trusted to do some thing well) | | | | | |
| 1.11 | **Health & Personality:** | (The various aspects that differentiate human from one another) | | | | | |
| 1.12 | **Initiative and Drive:** | (a self starter ability to action before being told what to do) | | | | | |
| 1.13 | **Sense of Responsibility:** | (Careful about assigned duties to deal with) | | | | | |
| 1.14 | **Flexibility & Adaptability:** | (ability to change or be changed easily according to the situation) | | | | | |
| 1.15 | **Stress Tolerance:** | (Handles pressure effectively without getting upset, moody) | | | | | |
| **2** | **TEACHING LEARNING PROCESS** | | | | | | |
| 2.1 | **Proficiency in teaching:** | (Training and practices in teaching ) | | | | | |
| 2.2 | **Personal Interest in Teaching:** | (The impact of ones interest in teaching profession) | | | | | |
| 2.3 | **Presentation & Communications skills**: | (The abilities of expression & interactions) | | | | | |
| 2.4 | **Speaking Style & Body language:** | (The impact of Communication through Gesture & poses) | | | | | |
| 2.5 | **Content knowledge:** | (The standard of knowledge delivered to students) | | | | | |
| 2.6 | **Lecture preparation:** | (The importance of lecturer preparation) | | | | | |
| 2.7 | **Language command:** | (The affects of the language that is used for teaching in class) | | | | | |
| 2.8 | **Response to Student queries:** | (The answer to student's questions during class) | | | | | |
| 2.9 | **Question Tackling:** | (The way questions or problem is tackled) | | | | | |
| 2.10 | **Courses taught:** | (B.Sc, Msc, Mphil, Phd, Post Doc) | | | | | |
| 2.11 | **Students Performance:** | (The percentage or standard of students results) | | | | | |
| 2.12 | **Work load:** | (The impact of work load per day on teaching) | | | | | |
| 2.14 | **Fairness in marking:** | (The accuracy of giving marks to students) | | | | | |
| **3** | **RESPONSIBILITY AND PUNCTUALITY** | | | | | | |
| 3.1 | **Punctuality:** | (Keeping the appointed time/ arrival to class & leave in time) | | | | | |
| 3.2 | **Checking Assignments in time:** | (The importance of in time checking) | | | | | |
| 3.3 | **Solving problems that pending in previous class:** | (Clarified pending class work) | | | | | |
| 3.4 | **Motivate the students in extra activities:** aside from curriculum) | (Guide students to participates in other activities | | | | | |
| 3.5 | **Willingness to work & seriousness to duty:** | (Self willing to do work & way to attempt) | | | | | |

|  |  | **ATTRIBUTES** | **LEVELS** |||||
|---|---|---|---|---|---|---|---|
|  |  |  | 5 | 4 | 3 | 2 | 1 |
| 3.6 |  | **Work Dedication:** (The ability to work hard & considered it is important) |  |  |  |  |  |
| **4** |  | **ADMINISTRATIVE SKILLS** |  |  |  |  |  |
| 4.1 |  | **Leadership:** (The ability to provides direction and motivates others to work for a common goal) |  |  |  |  |  |
| 4.2 |  | **Behavior when under pressure:** (The quality of work in a pressure situation) |  |  |  |  |  |
| 4.3 |  | **Judgments:** (The ability of Making connections between seemingly unrelated pieces of information, and understands ramifications of outcomes) |  |  |  |  |  |
| 4.4 |  | **Decision Making skills:** (The ability to take decision) |  |  |  |  |  |
| 4.5 |  | **Strategic Vision & Policy making skills**: (To Sets expectations and institution direction to meet goals and making strategies) |  |  |  |  |  |
| 4.6 |  | **Care of Rules & Regulation:** (The impact of following prescribed rules of working in organization) |  |  |  |  |  |
| 4.7 |  | **Controlling crises situation & uncertainty:** (The impact of uncertain or unpredicted situation) |  |  |  |  |  |
| 4.8 |  | **Listening suggestions of others:** (The willingness to gain idea by listening thinking of others) |  |  |  |  |  |
| 4.9 |  | **Taking Advantage from experience of others:** (The utilization of some one experience efficiently for own work) |  |  |  |  |  |
| 4.10 |  | **Ability to convince & motivate others:** (The Ability to reason and tracks of motivating others) |  |  |  |  |  |
| **5** |  | **SUPERVISOIN** |  |  |  |  |  |
| 5.1 |  | **Controlling students in class:** (The methods of handling and controlling students in class room) |  |  |  |  |  |
| 5.2 |  | **Students supervision** (B.Sc, Msc, Mphil, Phd, Post Doc) |  |  |  |  |  |
| 5.3 |  | **Supervision activities other than teaching**: (The quality of dealing others activities besides teaching) |  |  |  |  |  |
| 5.4 |  | **Interpersonal Relationships:** (The impact of social links with others) |  |  |  |  |  |
| **6** |  | **PROFESSIONAL ETHICS** |  |  |  |  |  |
| 6.1 |  | **Temperament and manners :** (The impact of nature and character of a teacher on performance) |  |  |  |  |  |
| 6.2 |  | **Interaction with students:** (The impact of communication & dealing with students) |  |  |  |  |  |
| 6.3 |  | **Interaction with Colleagues:** (The impact of communication & dealing with colleagues or co workers) |  |  |  |  |  |
| 6.4 |  | **Interaction with Officers:** (The impact of communication & dealing with officer or high authority personnel) |  |  |  |  |  |
| 6.5 |  | **Interaction with lower staff:** (The impact of communication & dealing with lower staff) |  |  |  |  |  |
| 6.6 |  | **Interaction with visitors/guests:** (The dealing with visitors or guest) |  |  |  |  |  |
| **7** |  | **RESEARCH ORIENTATION** |  |  |  |  |  |
| 7.1 |  | **Academic Class standing:** (The impact of ones strength in academics) |  |  |  |  |  |
| 7.2 |  | **Research potential:** (The affects of capability to conduct research) |  |  |  |  |  |
| 7.3 |  | **Standard of Projects:** (The successfulness or level of ones special works) |  |  |  |  |  |
| 7.4 |  | **Participations & organization of workshop, seminars, and conferences**: (interest in events) |  |  |  |  |  |
| 7.5 |  | **Research Production:** (The affects of the outcome of the research on teaching performance) |  |  |  |  |  |
| 7.6 |  | **Membership in research societies:** (The participation or involvement in research conducting bodies) |  |  |  |  |  |
| **8** |  | **PUBLICATIONS** |  |  |  |  |  |
| 8.1 |  | **Standard of Publications:** (The quality of published material) |  |  |  |  |  |
| 8.2 |  | **National/Foreign Journal Paper Publications:** (The interest or focus on journal papers) |  |  |  |  |  |
| 8.3 |  | **Joint Research Publications:** (The collaboration research work other national of foreign partners) |  |  |  |  |  |
| 8.4 |  | **Books/ Monographs Published** (College level, university level) |  |  |  |  |  |
| **9** |  | **AWARDS & ACHIEVEMENTS** |  |  |  |  |  |
| 9.1 |  | **Recognition:** (The impact self recognition in the institution) |  |  |  |  |  |
| 9.2 |  | **National/international awards**: (The effects of achieving national or international awards or prizes) |  |  |  |  |  |
| 9.3 |  | **Scholarships:** (The effects of achieving scholarships from national or foreign country) |  |  |  |  |  |
| 9.4 |  | **Research Grant received from Government/Private donors:** (The research grant in hands) |  |  |  |  |  |
| **10** |  | **COMPENSATION & REWARDS** |  |  |  |  |  |
| 10.1 |  | **Personal growth & Advancement** |  |  |  |  |  |
| 10.2 |  | **Presence of attractive compensation system** |  |  |  |  |  |

|  |  | ATTRIBUTES | LEVELS ||||| 
|---|---|---|---|---|---|---|---|
|  |  |  | 5 | 4 | 3 | 2 | 1 |
|  | 10.3 | **Presence of equitable internal salary** |  |  |  |  |  |
|  | 10.4 | **Presence of salary that reflects performance** |  |  |  |  |  |
|  | 10.5 | **Presence of salary that encourages better performance** |  |  |  |  |  |
|  | 10.6 | **Presence of salary that reflects standard of living** |  |  |  |  |  |
|  | **11** | **PROMOTION FACTORS** |  |  |  |  |  |
|  | 11.1 | **Presence of written and operational promotion policy** |  |  |  |  |  |
|  | 11.2 | **Provision of priority to seniority in promotion decision** |  |  |  |  |  |
|  | 11.3 | **Provision of priority to merit in promotion** |  |  |  |  |  |
|  | **12** | **JOB SECURITY & ENVIRONMENT FACTORS** |  |  |  |  |  |
|  | 12.1 | **work environment:** (Overall working environments in the work place) |  |  |  |  |  |
|  | 12.2 | **Highly secure job policy:** (The impact of secure job future, sure by organization or government ) |  |  |  |  |  |
|  | 12.3 | **cooperation from superiors:** (The support and behavior from high authority) |  |  |  |  |  |
|  | 12.4 | **Teamwork with colleagues:** (The Ability to works in groups and is a good team player) |  |  |  |  |  |
|  | 12.5 | **Security & Status:** (Overall security impact and social position in institute) |  |  |  |  |  |
|  | **13** | **ORGANIZATION EVALUATION POLICY** |  |  |  |  |  |
|  | 13.1 | **Presence of written and operational performance evaluation** |  |  |  |  |  |
|  | 13.2 | **Performance evaluation has a lot to do with salary** |  |  |  |  |  |
|  | 13.3 | **Performance evaluation has a lot to do with one's personal decisions** |  |  |  |  |  |
|  | 13.4 | **Provision of feed back of performance evaluation results** |  |  |  |  |  |
|  | 13.5 | **Performance evaluation is considered important task by superiors** |  |  |  |  |  |
|  | 13.6 | **Performance evaluation is knowledgeable** |  |  |  |  |  |
|  | **14** | **NEEDS & REQUIREMENTS** |  |  |  |  |  |
|  | 14.1 | **Psychological needs:** (The needs of breathing, Food, water, sleep, sex, homeostasis of human) |  |  |  |  |  |
|  | 14.2 | **Safety needs:** (The needs of security of body, employment, resources, morality, family, health and of property) |  |  |  |  |  |
|  | 14.3 | **Belong-ness needs:** (The needs of family, friends, life partner) |  |  |  |  |  |
|  | 14.4 | **Esteem needs:** (The needs of self esteem, confidence, respects to others, respect by others) |  |  |  |  |  |
|  | 14.5 | **Self-actualization needs:** (The needs of morality, spontaneity, and acceptance of facts) |  |  |  |  |  |
|  | **15** | **BACKGROUND FACTORS** |  |  |  |  |  |
|  | 15.1 | **Age:** |  |  |  |  |  |
|  | 15.2 | **Gender:** |  |  |  |  |  |
|  | 15.3 | **Qualification:** |  |  |  |  |  |
|  | 15.4 | **work Experience:** |  |  |  |  |  |
|  | 15.4 | **Religious Belief:** (The liking or affection of Religion) |  |  |  |  |  |
|  | 15.5 | **Political Inclination:** (The liking or affection of Politics) |  |  |  |  |  |